\documentclass[conference]{IEEEtran}
\IEEEoverridecommandlockouts
\usepackage{cite}
\usepackage{amsmath,amssymb,amsfonts}
\usepackage{algorithmic}
\usepackage{graphicx}
\usepackage{textcomp}
\usepackage{xcolor}
\usepackage{adjustbox}
\usepackage[numbers]{natbib}
\usepackage{subcaption}
\usepackage{multirow}
\usepackage{url}
\def\BibTeX{{\rm B\kern-.05em{\sc i\kern-.025em b}\kern-.08em
    T\kern-.1667em\lower.7ex\hbox{E}\kern-.125emX}}
\begin{document}

\title{Rotation, Translation, and Cropping \\for Zero-Shot Generalization}

\author{\IEEEauthorblockN{Chang Ye}
\IEEEauthorblockA{\textit{Game Innovation Lab} \\
\textit{New York University}\\
Brooklyn, USA \\
c.ye@nyu.edu}
\and
\IEEEauthorblockN{Ahmed Khalifa}
\IEEEauthorblockA{\textit{Game Innovation Lab} \\
\textit{New York University}\\
Brooklyn, USA \\
ahmed@akhalifa.com}
\and
\IEEEauthorblockN{Philip Bontrager}
\IEEEauthorblockA{\textit{Game Innovation Lab} \\
\textit{New York University}\\
Brooklyn, USA \\
philipjb@nyu.edu}
\and
\IEEEauthorblockN{Julian Togelius}
\IEEEauthorblockA{\textit{Game Innovation Lab} \\
\textit{New York University}\\
Brooklyn, USA \\
julian@togelius.com}
}

\IEEEpubid{\begin{minipage}{\textwidth}\ \\[12pt]
978-1-7281-4533-4/20/\$31.00 \copyright 2020 IEEE \end{minipage}}

\maketitle

\begin{abstract}
Deep Reinforcement Learning (DRL) has shown impressive performance on domains with visual inputs, in particular various games. However, the agent is usually trained on a fixed environment, e.g. a fixed number of levels. A growing mass of evidence suggests that these trained models fail to generalize to even slight variations of the environments they were trained on. This paper advances the hypothesis that the lack of generalization is partly due to the input representation, and explores how rotation, cropping and translation could increase generality. We show that a cropped, translated and rotated observation can get better generalization on unseen levels of two-dimensional arcade games from the GVGAI framework. The generality of the agents is evaluated on both human-designed and procedurally generated levels. 
\end{abstract}


\section{Introduction}

The way in which a problem or data is represented has a large effect on how easy it is to be learned by a machine learning method. For example, it is common knowledge that when trying to learn features expressed as categorical variables, it makes all the difference in the world whether this is presented to the algorithm as a one-hot encoding or as different values of a single input. With a one-hot encoding, learning might work as intended, whereas the other encoding has much less chance of working.

This kind of knowledge, crucial as it may be, is commonly not the subject of a paper of its own, but introduced as a detail in papers focusing on some other method. For example, in the recent AlphaStar paper, one of the components was the use of a transformer architecture to process variable-length lists of units in the game~\cite{vinyals2019grandmaster}. %One exception to this is Thermometer Encoding, which was the subject of its own paper~\cite{buckman2018thermometer}. 

\begin{figure}
    \centering
    \includegraphics[width=0.6\linewidth]{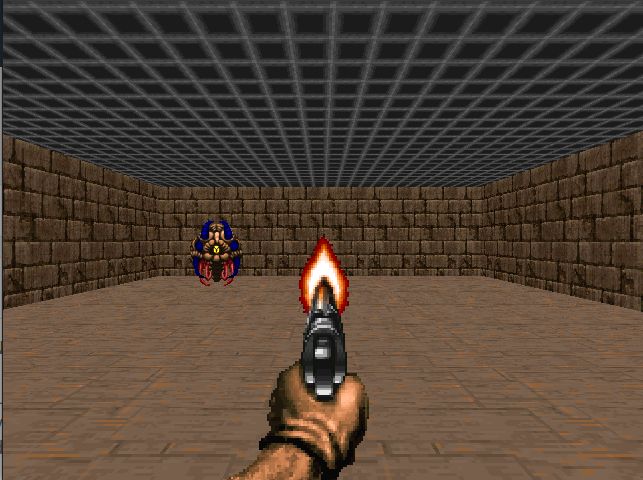}
    \caption{VizDoom’s first-person perspective.}
    \label{fig:vizdoom}
\end{figure}

The ability to learn straight from pixels, using the same information as people, is one of the reasons deep reinforcement learning has became so popular~\cite{mnih2015human}. Yet, learning from pixels poses a number of challenges.%A matrix of color values for pixels representing shapes optimized for the human eye cannot reasonably be expected to be optimal for a neural network. 
In some ways, these outputs are not obvious for humans either; a human that does not know how to play videogames need to first learn that they ``are'' the character that moves around which they can control with the joystick, and if they play a game where the player character moves non-holonomically (such as a char) they need to learn that a particular direction of the joystick means something different depending on which way the player agent is facing. These conventions are largely carried over between games, which explains why people can so rapidly pick up a new game and start playing. However, just observe someone who does not know the perception and control conventions of a game genre try to play a game of that genre, to understand how non-obvious the conventions are. Which makes it even more surprising that deep reinforcement learning agents can learn to play these games so well.

Over the last few years, several papers have started questioning what deep neural networks that learn to play from pixels actually learn. In ``Playing Atari with Six Neurons'', it is shown that surprisingly small networks with a few hundred parameters can learn to play many Atari games with a skill that rivals that of networks with hundreds of thousands, or even millions, of neurons~\cite{cuccu2019playing}. This is accomplished by separating out the preprocessing and learning a library of sensory prototypes, allowing for an input that encodes how similar a particular observation is to other observations. This work questions what a giant neural network actually does if the policy can be encoded by a tiny neural network; perhaps most of the network is engaged in some kind of simple transformation of the input image?

Another way to investigate what deep networks, trained with reinforcement learning, learn from pixel data is through studies of their generalization capacity~\cite{cobbe2019quantifying,song2020observational}. These studies generally have rather negative results. For example, a set of experiments showed that networks trained at one or a small set of levels could not play other levels of the same game they had been trained on ~\cite{justesen2018illuminating}. Training using procedural content generation, where each episode uses a new level, managed to create networks that generalized somewhat better, but not much better.

Anecdotal evidence suggests that this failure of generalization extends to various environments with a static third-person perspective\footnote{With a static third-person perspective, we mean one which does not change depending on the movement of the agent, or only does so rarely (for example, when moving between rooms in a flick-screen fashion).}. On the other hand, games such as Doom (shown in figure~\ref{fig:vizdoom}), which are seen from the vantage point of the agent, do not seem to create the same generalization problem for deep reinforcement learning.

This paper builds on the hypothesis that deep reinforcement learning cannot easily learn generalizable policies for games with static third-person views, but that they can do so when the same game is seen through a more agent-centric view. It tests this hypothesis by training deep networks with reinforcement learning on multiple different games, with and without various perceptual modifications, in particular rotation and translation. For each version, we report performance on a training set of levels and, separately, on a larger test set.

It should be pointed out that while the hypothesis advanced here might seem obvious and the experiments somewhat simplistic, the hypothesis runs counter to received wisdom and implicit assumptions in the mainstream of deep reinforcement learning research. We are saying that deep reinforcement learning on games with static third-person representations, in general, does not work in the sense that it does not learn generalizable policies. This may or may not be because these network structures cannot learn the types of input transformations that are necessary for generalizable policies. In any case, we imply that input representation plays a much larger role than is commonly assumed.

% ---

% background: drl on visual input from atari-like environments doesn't generalize (see illumination paper, openai research)

% however, it seems to generalize better on doom and similar environments

% hypothesis: it's because of the third-person static view. Neural nets can't, or won't if they don't have to, learn to translate and rotate to a first-person view.

% we investigate this...

\section{Background}

Deep Reinforcement Learning has a lot of success in video games, especially arcade video games like the atari environment \cite{mnih2015human}. But, as has been mentioned, these works primarily focused on learning to play mostly deterministic environments \cite{justesen2019deep}. With the goal of teaching agents a real understanding of an environment so that it can be robust and useful outside of simple video game situations, there has been a lot of research recently in improving the zero-shot and few-shot generalization ability of RL agents.

It isn't immediately obvious from the research that agents are bad at generalization. Results from agents such as AlphaStar \cite{vinyals2019grandmaster}, OpenAI's Hide and Seek \cite{Baker2020Emergent}, and agents from the Doom competition \cite{wydmuch2018vizdoom} all make it appear that agents can learn general policies that adapt very well to many situations. Without discussing how general these learned policies are, we note that there are key components in these environments that would help with generalization. Both Starcraft and Hide and Seek are environments that allow self-play between competing teams. This provides a natural curriculum for an agent to learn a robust and general policy. The other component is the observation space provided in these environments. In VizDoom and Hide and Seek the agent is shown an agent-centric view of the world, making it easier to see how the agents' actions affect the world. Hide and Seek also provides extra information about the global world state. For AlphaStar, the agent is shown a minimap of the entire game but it is also provided key summary information about the objects in the game world, which allow the agent to once again have immediate feedback on how its actions are affecting the environment.

Attention has only recently been growing over the difficulty most game environments provide for learning general policies. To combat this, a number of new environments and benchmarks have been released to provide test grounds for how easily each algorithm can learn a general policy. Of the many that have been introduced, several big ones are; Coin Run~\cite{cobbe2019quantifying}, Obstacle Tower~\cite{juliani2019obstacle}, General Video Game AI (GVGAI)~\cite{torrado2018deep}, and Maze Explorer ~\cite{harries2019mazeexplorer}. These environments focus on having lots of games and levels in order to provide the training data for an agent to learn a general policy. With a large number of different environments to train on, the agent cannot simply memorize a sequence of actions to take for every environment. These frameworks are very useful but agents do not automatically learn general policies even in the presence of unlimited new levels for a simple game as found by ~\cite{justesen2018illuminating}. While there have been some promising results on these environments, it must be noted that this is not a solved problem.

To make matters worse, there are many situations where it is impossible to generate a large number of different versions of an environment for training, so it is important for agents to learn as general a policy as possible from a small set of environments. To help with this, researchers have found ways to inject noise into the training process. Even in the original Atari Deep Q-Learning work, they would have the agents take a random number of noop (no action) steps at the beginning of a game to randomize the initial layout of the game \cite{mnih2015human}. This would prevent the agent from simply memorizing a sequence of actions without reacting to the environment. Another simple approach to increase noise during training is sticky actions \cite{machado2018revisiting}. Sticky actions introduce a parameter which is the probability that an agent's action will be repeated instead of a new action is being calculated. This introduces randomness into training and forces the agent to learn a policy that is not too brittle. More recently \cite{igl2019generalization} experimented with smarter ways to introduce noise for the agent's actions. Instead of injecting noise when an agent is taking an action, which results in the agent collecting worse data, the agent collects data and noise is added during the network update. This helps the agent learn an improved policy but it still relies on lots of training levels.

Focusing more on the visual aspect, \cite{zhang2018natural} proposes adding visual noise into the game environment. This requires them to know which part of the frame is from the background and which part is from objects. They then replace the background with either gaussian noise or video frames from the natural world. With this noise, the agent's performance plummeted showing that they could not learn a general policy that ignored the background area of the screen. To further the understanding around this, \cite{song2020observational} examined how pixels in a state observation can provide unnecessary information that agents use to memorize a brittle policy.

Recent work that has come out after this work has focused on data augmentation, in particular image translation, as a generalization technique. In two consecutive works, researchers proposed using random augmented observations both with and without additional loss functions during training \cite{laskin2020reinforcement,srinivas2020curl}. The simple techniques they used achieved the state-of-the-art performance on the majority environments on the DeepMind Control benchmarks. Kostrikov et al  proposes using augmented observations to reduce the variance of Q-function estimation, so as to stabilize the training process
~\cite{kostrikov2020image}. Specifically, they apply a set of data augmentation techniques $K$ times, and then calculate an average $Q$ value estimation to reduce the variance. The success of these approaches along with our results suggest there is a connection between an agent's point-of-view and that point-of-view's data-augmentation characteristics.

%To further this direction of research, we focus on ways to augment the observation space of the environment to assist agents in learning general policies.

%six neurons
%\cite{cuccu2019playing}

%old work in evolutionary robotics learning general behavior with simple lo-fi egocentric sensors, also work on evolving car racing skills

%convolutional nets: what can they actually learn? locality invariance is not the same as translation

\section{Generalization Approach}
%\section{Change in Perspective}

% In this work, we evaluate how the agent's point of view impacts how and what it learns during deep reinforcement learning. One difficulty in reinforcement learning is for the agent to understand the consequences of its actions. An agent takes an action and the state changes, some changes are caused by the agent and others would happen anyway. In environments where the agent is embodied in the environment and shown a map of the entire world, it really struggles with learning from a third-person point of view \cite{justesen2018illuminating}. It has to learn that it's actions control a small blob of pixels in the environment and that it controls what that blob does to other things around it all the while the entire environment is changing everywhere on its own.

In this work, we are looking at improving the agent's generalization through modifying the input representation and with no data augmentation or transfer learning. We are taking a step toward a better understanding of the optimal representation for reinforcement learning. In environments where the agent is embodied in the environment and shown a map of the entire world, it really struggles with learning from a static third-person point of view \cite{justesen2018illuminating}. The network not only needs to learn the consequences of its actions but also needs to track a small blob of pixels to know its location. We propose to transform the input representation to be more centered around the agent as it is seeing it from its point of view. Doing that will reduce the number of tasks the agent needs to learn during training.

We propose always giving the agent an agent-centric\footnote{Note that ``agent-centric'' is not the same as ``first-person''. With an agent-centric view, we mean any view that puts the representation of the agent (a.k.a. its avatar) at the center; in the examples here, we use a third-person agent-centric view.} view when possible and further propose that cropping the agent’s view to just its immediate surroundings can greatly improve its ability to learn in Deep Reinforcement Learning (DRL). We propose three techniques to do this that can be applied to any environment with a visible agent even if the only state information available is a pixel image. We propose rotating, translating, and cropping the observation around the agent’s avatar. These are quick transformations that can be applied to the observation image and they only require knowledge of the location and the direction of the agent’s avatar. If the location information isn’t available from the environment, a simple object detection algorithm can be used to find the avatar image on the screen. For the avatar's direction, if relevant to the environment, it can be extracted from the agent's actions as the agent usually need to change its direction before it starts changing its location. Rotation keeps the agents always facing forward, so any action it takes always happens in the same relative direction to it. Next, translation translates the observation around the agent so its always in the center of its view. Finally, cropping shrinks the observation down to just local information around the avatar.

These changes at first can appear like obvious transformations, but we did not find anywhere in the literature discussing how observation perspective affects learning for DRL. We not only recommend these techniques for people working with map-like views, but also measure their effectiveness and discuss where and when they are useful. A local, agent-centric view, allows for better learning in our experiments and the policies learned generalize much better to new environments even when trained on only five environments. This implies the agent was able to learn from the correct objects in the environment instead of just memorizing states from pixels.

In practice we also found it was necessary to randomize the agent’s initial orientation, and to replace the agent’s avatar with a square. We believe the randomization is necessary to stop the agent from memorizing an opening sequence. We believe the agent was using the transformed avatar’s layout to determine its global orientation, but also found that the agent performed better in every single experiment if the avatar was replaced.

In the following subsections, we are going to explain in detail these three transformations. These transformations can be used by themselves or combined together to combine their effects.

% 1. Translation

% What is translation? We pad the input and crop it such that the avatar is always at the center, and the entire window is visible.

% Theory: this makes the view ego-centric. The agent sees exactly how it's actions affect the world. It does not need to learn to position itself. It can learn more general rules as actions have more consistent, generic results. As opposed to movmement resulting in changes all over the screen that the agent has to learn are all the same action.

% 2. Rotation

% What is Rotation? - Describe how it works

% Theory: Action always affects the same local area...

% 3. Cropping

% What is Cropping? - ...

% Theory: Learn local rules, data augmentation, most arcade games are heavily biased towards local effects

% To apply our transformations, we need to know the player's avatar location and direction. In most of the games, this information is not available directly to the from the pixel data but we can use a simple OpenCV image tracking function using the player picture to detect that position. The avatar direction, if relevant to the environment, can be extracted from the agent's actions as the agent usually need to change direction before it starts changing its location.

% We are proposing three different transformation techniques: Translation, Rotation, and Cropping as ways to battle the lack of generalization in the reinforcement learning. We will discuss these techniques in the following subsections. These techniques can be used by themselves or combined together to combine their effect.

\subsection{Translation}

\begin{figure}[ht]
    \centering
    \includegraphics[width=.9\linewidth]{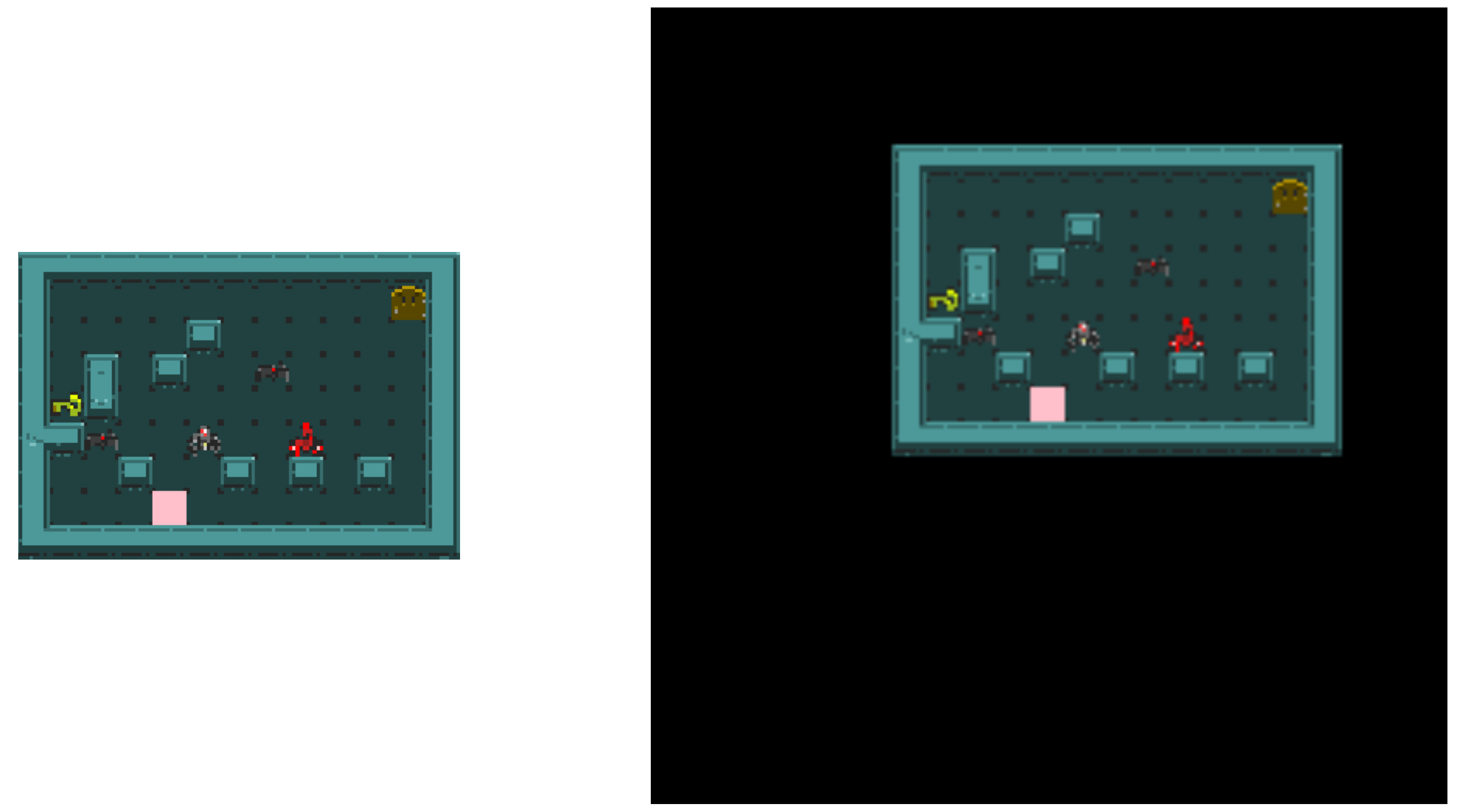}
    \caption{The left observation is being translated around the player's avatar (pink rectangle) to the right observation.}
    \label{fig:translation}
\end{figure}

Translation is the process of centering the observation image around the player's avatar. The idea is the player's avatar should always be in the center of the observation image after this transformation. Figure~\ref{fig:translation} shows the translation process where the observation appearing on the left was padded with black pixels and centered around the player's avatar (pink rectangle). We can see that the center pixel of the new observation is the player's avatar. Translation will restrict the avatar to learn the relative position to other game objects such as a key, door, or enemy in figure~\ref{fig:translation} \cite{kwok2004reinforcement} which is useful for the agent to understand the game and take the corresponding action. In a lot of video games, having a relative position is enough to win the game as you usually need to move the avatar toward a certain target to interact with it.

\subsection{Rotation}

\begin{figure}[ht]
    \centering
    \includegraphics[width=.9\linewidth]{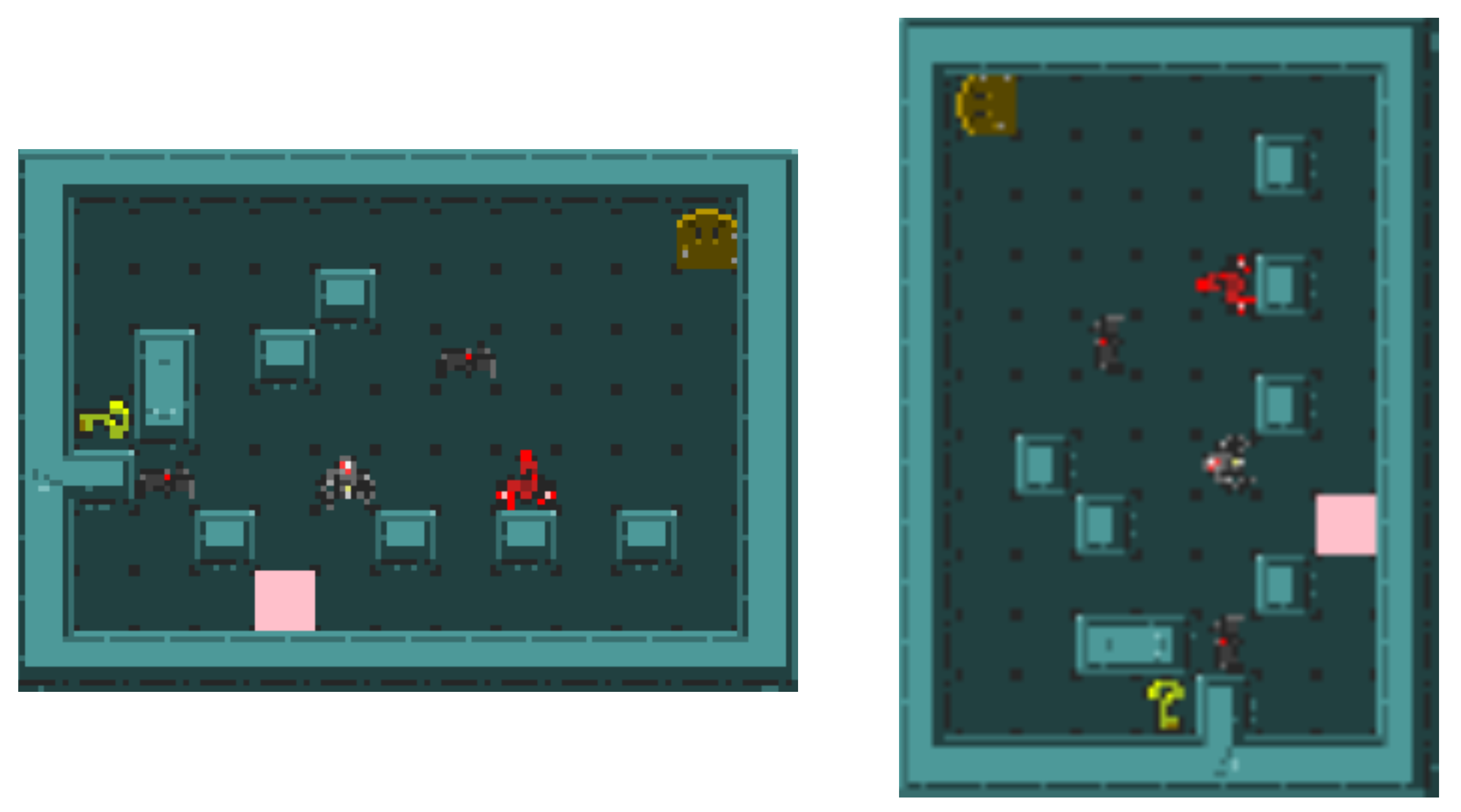}
    \caption{The left observation is being rotated using the player's avatar (pink rectangle). In the original the player is looking right, the observation is rotated so that the agent is always looking up.}
    \label{fig:rotation}
\end{figure}

Rotation is the process of orienting the observation to face the same direction as the player's avatar. Figure~\ref{fig:rotation} shows the rotation transformation on the left observation (the original observation). The rotated observation shown on the right is the original observation rotated towards the player direction (which is right in that state). An important note when using rotation transformation, the action space the agent is taking has to be unrotated before it is returned to the framework. For example: in figure~\ref{fig:rotation} the avatar is facing right so the observation was rotated by $90^{\circ}$. If the avatar wants to move up, in the new observation this is technically right in the original framework. To solve that problem, any action happening that is taking place in the environment has to be rotated in the negative direction of the rotation degree. Rotation helps the agent to learn navigation as it simplifies the task. For example: if you want to reach for something on the right, the agent just rotates until that object is above and then moves up. If that object becomes to the left, the same strategy can be applied (rotate then move up). This is not the case without rotation where we need to move in a different direction depending on the location of the target object.

% The direction avatar is facing is always the action in the last step. We save the previous action in each step and rotate the observation by a certain degree based on the angle between the previous action and direction "UP".
% \par It's also important to notice that the action agent is taking is not the real action in the environment as the direction in a rotated observation may not be the same direction in the original observation. We use a simple degree system to recover the real action. Since the action except None and Use has 4 directions which are up, left, down and right. We represent actions with digits $0,1,2,3$. Each digit represents the number of 90 degrees it should take from direction up. Note that we are using counter-clockwise rotation in this experiment. The real action could be expressed as the following equation.
% \begin{align}
%     \textit{real action} = f((f^{-1}(action)+4-d(direction))%4)
% \end{align}
% where $f$ is a function mapping the digit to action, $f^{-1}$ is the inverse function of $f$. $d$ is a function mapping direction to the digit.

\subsection{Cropping}

\begin{figure}[ht]
    \centering
    \includegraphics[width=.9\linewidth]{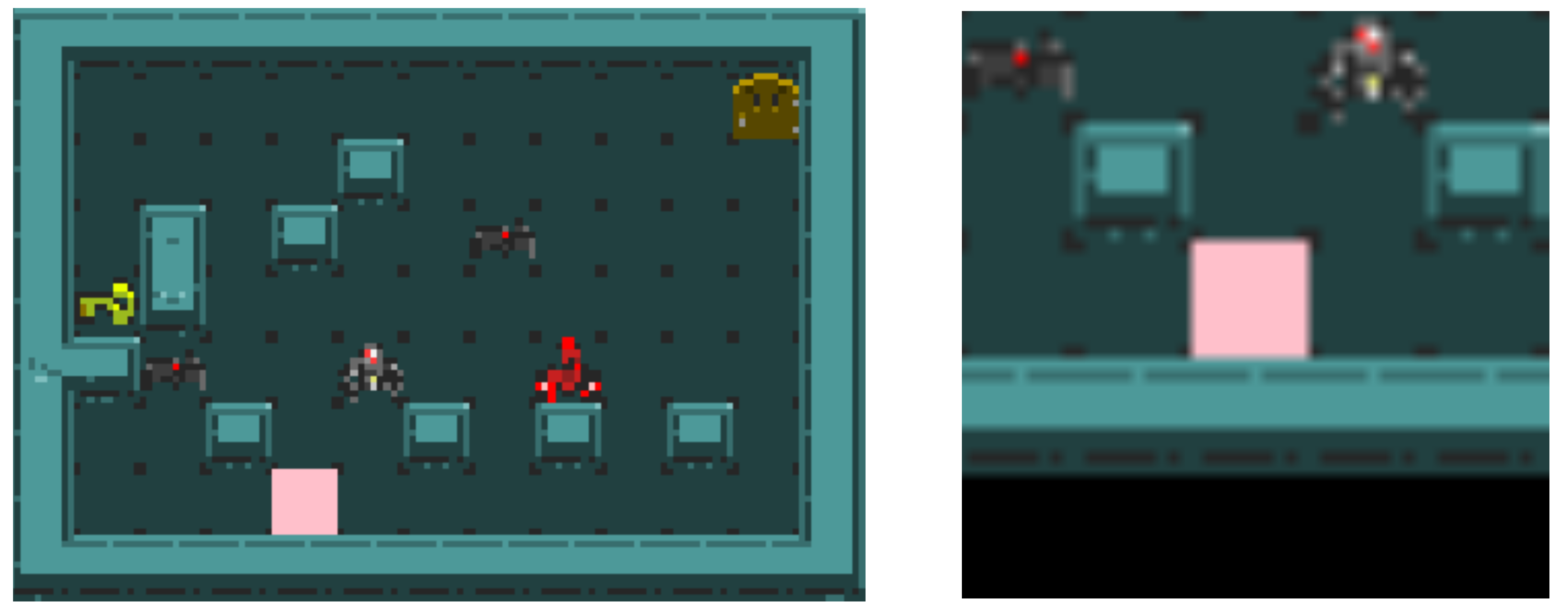}
    \caption{The left observation is being cropped using the player's avatar (pink rectangle) position to the right observation.}
    \label{fig:cropping}
\end{figure}

Cropping is the process of only showing the observation around the player and not the full observation. Cropping by default is a translation technique as the new observation is centered around the avatar. Figure~\ref{fig:cropping} shows a 5x5 cropping transformation being applied to the left observation (the original game observation) to a smaller view that is centered around the player (the center pixel is the center of the avatar). 

The cropping helps to reduce the state space of what the agent is seeing to a smaller subset which can help the agent to learn a generalized policy. Neural Nets can be interpreted as behaving as Locality-sensitive hashing (LSH) functions and they intelligently learn to recognize similar states. Larger environments with many combinatorial arrangments of agent and object locations make it more difficult for the agent to understand what states are functionally the same. In a cropped view the agent can directly see the effect of its action, assuming its actions are local, and it can directly match actions to states.

In many video games actions and interactions mostly happen locally, cropping focuses the agent to this area. This helps in many games but obviously also provides a disability in the form of missing information and lack of a global context. Another reason is that cropping observation could be considered as a data augmentation which is helpful for generalization as it learns a broader set of state-action values, also referred to as the Q-values \cite{watkins1992q}.

\section{Experimental Methods}

% gvgai

% zelda game

% standard rl algorithms setup

% how we do translation and rotation

% blacking out the avatar

% network size (we should probably try several different ones)

\begin{figure*}[ht!]
     \centering
     \begin{subfigure}[t]{0.19\linewidth}
         \centering\captionsetup{width=.9\linewidth}
         \includegraphics[height=2cm]{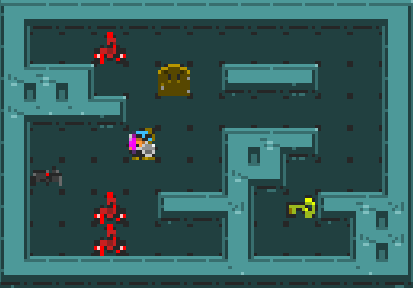}
         \caption{human-designed zelda level}
         \label{fig:human_level}
     \end{subfigure}
     \begin{subfigure}[t]{0.19\linewidth}
         \centering\captionsetup{width=.9\linewidth}
         \includegraphics[height=2cm]{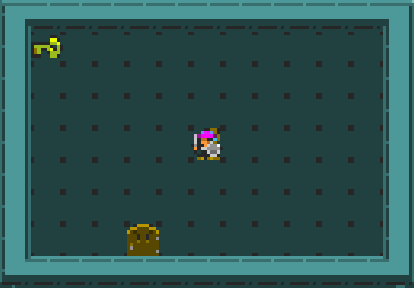}
         \caption{simple zelda training level (key and door always on the left)}
         \label{fig:obs_simple_train}
     \end{subfigure}
     \begin{subfigure}[t]{0.19\linewidth}
         \centering\captionsetup{width=.9\linewidth}
         \includegraphics[height=2cm]{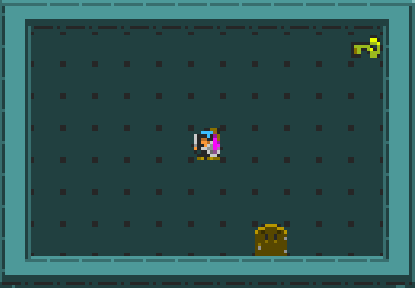}
         \caption{simple zelda test level (key and door always on the right)}
         \label{fig:obs_simple_test}
     \end{subfigure}
     \begin{subfigure}[t]{0.19\linewidth}
         \centering\captionsetup{width=.9\linewidth}
         \includegraphics[height=2cm]{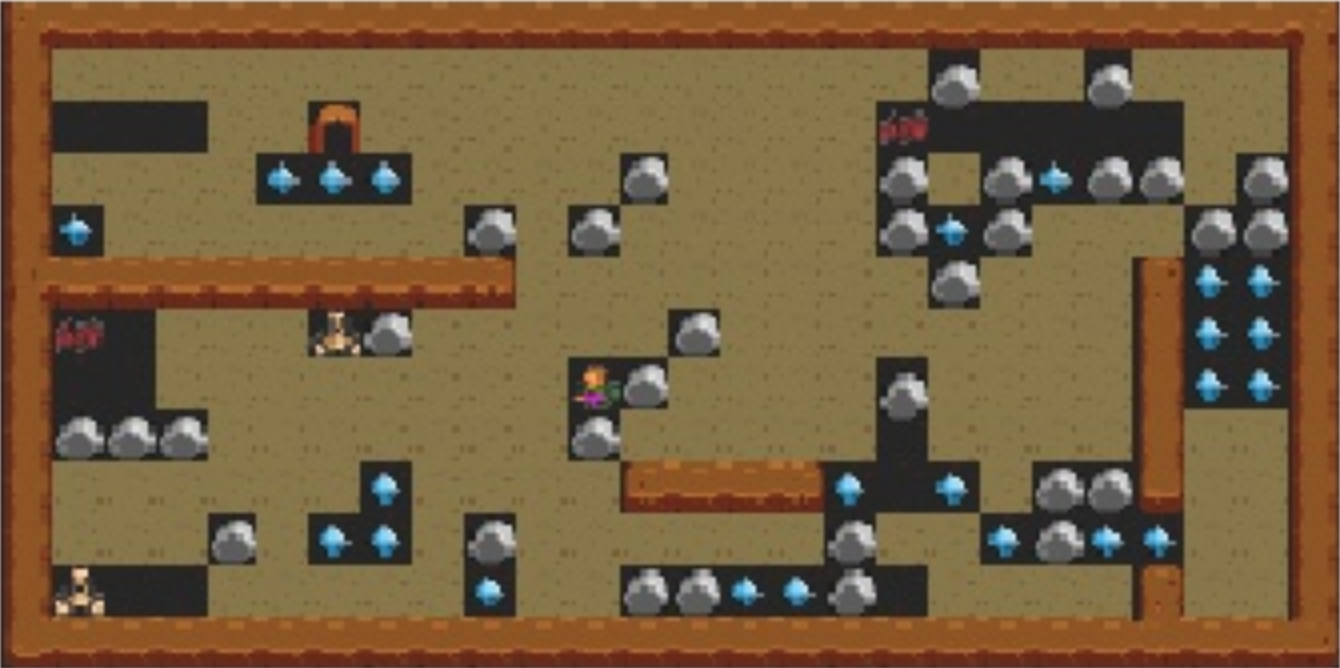}
         \caption{human-designed boulderdash level}
         \label{fig:human_boulderdash}
     \end{subfigure}
     \begin{subfigure}[t]{0.19\linewidth}
         \centering\captionsetup{width=.9\linewidth}
         \includegraphics[height=2cm]{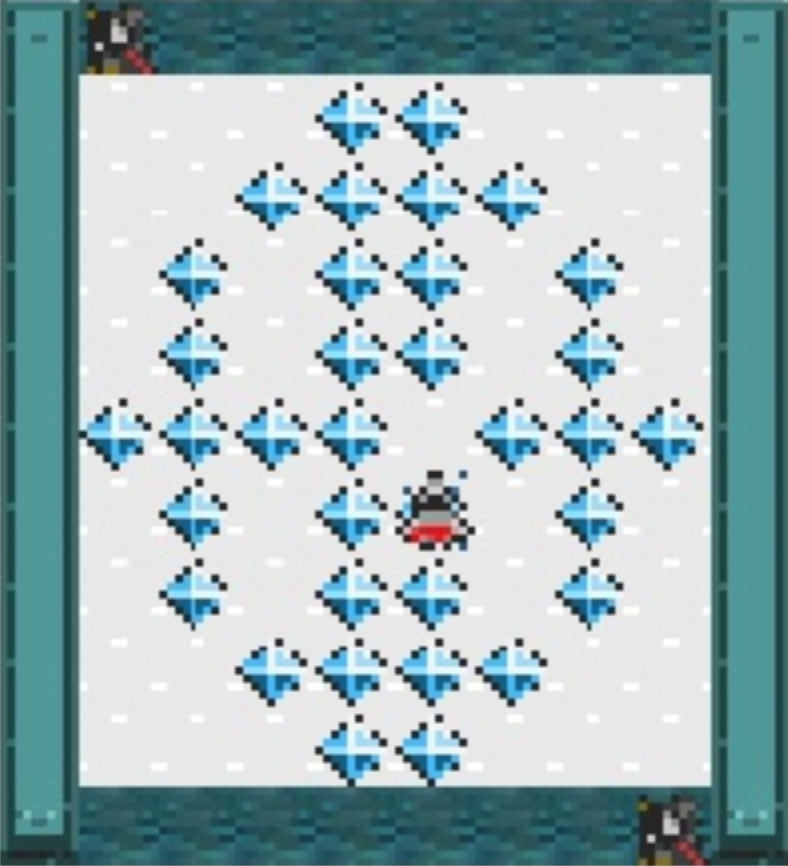}
         \caption{human-designed solarfox level}
         \label{fig:human_solarfox}
     \end{subfigure}
     \caption{Examples of game levels from zelda, simple zelda, boulderdash, and solarfox.}
     \label{fig:train_test_levels}
\end{figure*}

In this work, we use the OpenAI Gym~\cite{openai2016brockman} interface of the GVGAI Framework~\cite{torrado2018deep}. We test our techniques on three different games and one game variant:
\begin{itemize}
    \item \textbf{Zelda:} is a GVGAI port for the dungeon system in The Legend of Zelda (Nintendo, 1986). The goal of the game is to get a key and reach the exit door while avoiding hitting enemies. The agent also can use its sword to kill enemies for additional scores. Figure~\ref{fig:human_level} shows an example of a human-designed Zelda level.
    \item \textbf{Simple Zelda:} similar to the Zelda game but it only has a key and the door. The agent's goal is to get the key and reach the door. In this game, there are no walls so the agents don’t need to learn navigation. It just needs to learn the goal of the game. Also, all the game levels are designed such that the player starts in the center of the map and the key and door are both either on the left of the agent or the right shown in figures~\ref{fig:obs_simple_train} and \ref{fig:obs_simple_test}.
    \item \textbf{Boulderdash:} is a GVGAI port for Boulder Dash (Data East, 1984). The goal is to collect 10 different diamonds then reach the goal while avoiding getting killed by enemies or the falling boulders. Figure~\ref{fig:human_boulderdash} shows an example of one of the training levels in Boulder Dash.
    \item \textbf{Discrete Solarfox:} is an adapted version of a GVGAI port for Solarfox (Midway Games, 1981). The goal of the game is to collect all the diamonds without hitting the borders of the map or enemy bullets. A complication is that the avatar is always moving; if no new input is given, it keeps moving in the same direction as the last frame. We modified this game by increasing the avatar speed by factor of $7$ as the framework only returns the avatar location in integer values (while actual speed was $1/7$ in the original game). Figure~\ref{fig:human_solarfox} shows an example of one of the training levels in Solarfox where the player controls the spaceship.
\end{itemize}

To evaluate our methods. We employ the \textit{Advantage Actor-Critic} (A2C) algorithm. Specifically the implementation from Open AI Baselines~\cite{dhariwal2017openai}. The neural network has the same structure as in \citeauthor{mnih2016asynchronous}~\cite{mnih2016asynchronous} with a body consisting of three convolutional layers followed by a single fully-connect layer. We trained the agents until convergence (which took between 200 million frames to 400 million frames). We configured the A2C algorithm to use a step size 5, 84 by 84 wrapped frame, 4 frames per stack and a constant learning rate of 0.007 with the RMSProp optimizer.

\begin{figure}[ht]
     \centering
     \begin{subfigure}[t]{0.45\linewidth}
         \centering
         \includegraphics[width=\linewidth]{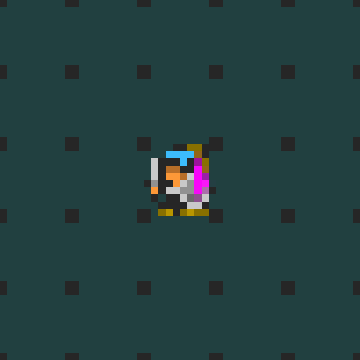}
         \caption{Original view}
         \label{fig:normal}
     \end{subfigure}
     \begin{subfigure}[t]{0.45\linewidth}
         \centering
         \includegraphics[width=\linewidth]{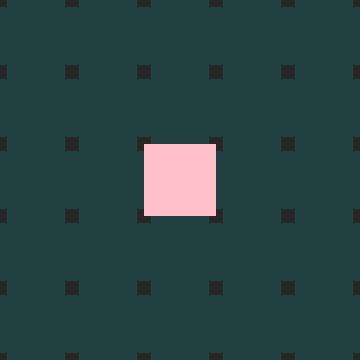}
         \caption{View with replacing avatar}
         \label{fig:repava}
     \end{subfigure}
     \caption{Examples of observation with or without replacing avatar}
     \label{fig:obs_repava}
\end{figure}

In preliminary experiments, we found that agents trained on Simple Zelda levels barely won on the test set. When we investigated the playtraces we found that the agent is simply memorizing to go left (where the key and the door are in the training set) instead of understanding where the key and the door location are and trying to move toward them. This was still happening in the cropped view with rotation and translation, where the agent has no idea where it could find the key and the door and doesn't have an idea what is left (because of rotation). we discovered that the agent uses the avatar rotated pixels to memorize the solution for different levels similar to Song et al's work~\cite{song2020observational} where the agent used the scoreboard to solve the game. To avoid that problem, we simply replace the avatar with a square of a certain color shown in figure~\ref{fig:obs_repava}.

% During the experiments, the agent trained on the levels where the keys and doors are on the left, barely win the test levels where the keys and doors are on the right. We investigated play traces from the test levels and the play traces indicate that the agent simply memorizes the first action in the training phase as the first step in each training episode is always some fixed action. To avoid the overfitting problem, we simply replace the avatar with a square with some random color (see Figure \ref{fig:obs_repava}). By doing that. The agent will not overfit easily as there will be only one initial observation so that the probabilities of each action are almost the same. The policy entropies from the two experiments Figure \ref{fig:policy entropy} where the only difference is replacing avatar also proves this assumption.

% To verify our approach. We test it on a simple version of Zelda and human-designed Zelda environments. In the simple version of Zelda, the environment only has one door, one key and one avatar. The avatar is always placed at the center and key and door are placed at same side of avatar(see Figure \ref{fig:obs_simple_test} and \ref{fig:obs_simple_train}). The level set contains all the combinations of key and door locations on one side. The human-designed levels are collected from ****** (see \ref{fig:human_level})

For Zelda, we train the agent on the 5 human-designed levels and test it on a different 45 human-designed levels. While for Simple Zelda, we train the agent on levels where the key and the door are on the left side of avatar and test it on levels where the key and the door are on the right side of the avatar. The idea behind that is to test generality in its most simple form where the agent needs to understand where to go and not just memorize the sequence of actions to win the level. The number of levels in train and test set is the same which is 1190 levels which reflect all the possible levels where the key and the door are assigned on either side of the avatar. For Boulderdash and Solarfox, the agent is trained on the 5 human-designed levels that come with the framework and tested on 50 different generated levels using the generator from \citeauthor{justesen2018illuminating}'s work~\cite{justesen2018illuminating}. For all the experiments, the avatar starts with a random direction uniformally sampled from all four directions to allow the agent not to memorize the starting direction. Similar to the noop random initialization in \citeauthor{mnih2015human}~\cite{mnih2015human} work.

We trained 3 models for all the possible combinations of our proposed transformations (Translation, Rotation, and Cropping) on all the proposed problems (Simple Zelda, Zelda, Boulderdash, and Solarfox). We end up with having 6 total experiments instead of 8 because you can't do Cropping without Translation. For the avatar's location, we extract it from the game engine itself which will be replaced in future work with a simple OpenCV image tracking function.

% Can also incorporate other known generalization technques, such as random starts \cite{mnih2015human}. We use this.

% \begin{figure}
%     \centering
%     \includegraphics[width=0.8\linewidth]{images/newplot.png}
%     \caption{Policy Entropy}
%     \label{fig:policy entropy}
% \end{figure}
% \par

\section{Results}

%to Philip: need to visualize the results graphically - each approach and relative generalization - simple bar graph?
\begin{table*}
\centering
\resizebox{\linewidth}{!}{
\begin{tabular}{|c|c|c|c|c|c|c|c|c|c|c|}
\hline
\multirow{2}{*}{crop} & \multirow{2}{*}{translate} & \multirow{2}{*}{rotate} & \multicolumn{2}{c|}{simple zelda} & \multicolumn{2}{c|}{zelda} & \multicolumn{2}{c|}{boulderdash} & \multicolumn{2}{c|}{dsolarfox} \\ \cline{4-11} 
                  &                   &                   &           train & test & train & test & train & test & train & test \\ \hline
       0           &         0          &       0            &           $100.0 \pm 0.1\%$      &     $0.0 \pm 0.0\%$      &    $76.7\pm17.6\%$       &       $0.4 \pm 0.8\%$     & $19.0\pm29.1\%$ &  $0.0\pm0.0\%$ & $87.7 \pm 15.1\%$ & $49.0\pm7.1\%$ \\ \hline
           0       &        0           &        1           &           $100.0\pm0.0$\%       &    \textcolor{red}{$70.5\pm10.4$\%}       &      $81.0\pm14.8$\%     &      $1.1\pm1.3$\%         & $12.3\pm11.6$\% & $0.1\pm0.4$\% & $29.5\pm17.8\%$ & $1.5\pm1.5\%$ \\ \hline
           0       &        1       &           0        &           $100.0\pm0.1$\%      &      $3.8\pm1.6$\%     &    $78.7\pm18.6$\%      &     $1.0\pm1.5$\%      &  $30.0\pm12.6$\% & $0.2\pm0.5$\%  &  $87.0\pm14.5\%$  & $86.1\pm3.0\%$ \\ \hline
           0       &        1           &       1            &           $100.0\pm0.0\%$      &      $49.0\pm1.0\%$     &     $75.0\pm19.2\%$      &     $0.9\pm1.4\%$   &  $28.3\pm11.7\%$  &  $0.0.\pm0.3\%$ & $74.0\pm16.9\%$  & $55.1\pm12.2\%$ \\ \hline
           1       &        1           &       0            &          $100.0\pm0.0\%$       &    $14.9\pm3.1\%$       &    $66.7\pm21.8\%$       &     $5.2\pm2.8\%$      & $8.3\pm12.3\%$ & $0.8\pm1.3\%$ & $95.7\pm9.0\%$ & \textcolor{red}{$90.7\pm6.8\%$} \\ \hline
    1   &   1     &     1          &           $99.9\pm0.1\%$      &      $62.9\pm3.6\%$     &     $65.7\pm19.7\%$      &   \textcolor{red}{$22.0\pm4.5\%$}   & $10.7\pm12.4\%$  &   \textcolor{red}{$1.0\pm1.3\%$}   & $85.5\pm14.8\%$ & $86.4\pm3.8\%$ \\ \hline
\end{tabular}
}
\caption{Test and train result of different combination in simple zelda, zelda, boulderdash and discrete solarfox }
\label{table:results}
\end{table*}

\begin{figure*}
    \centering
     \begin{subfigure}[t]{0.24\linewidth}
         \includegraphics[width=\linewidth]{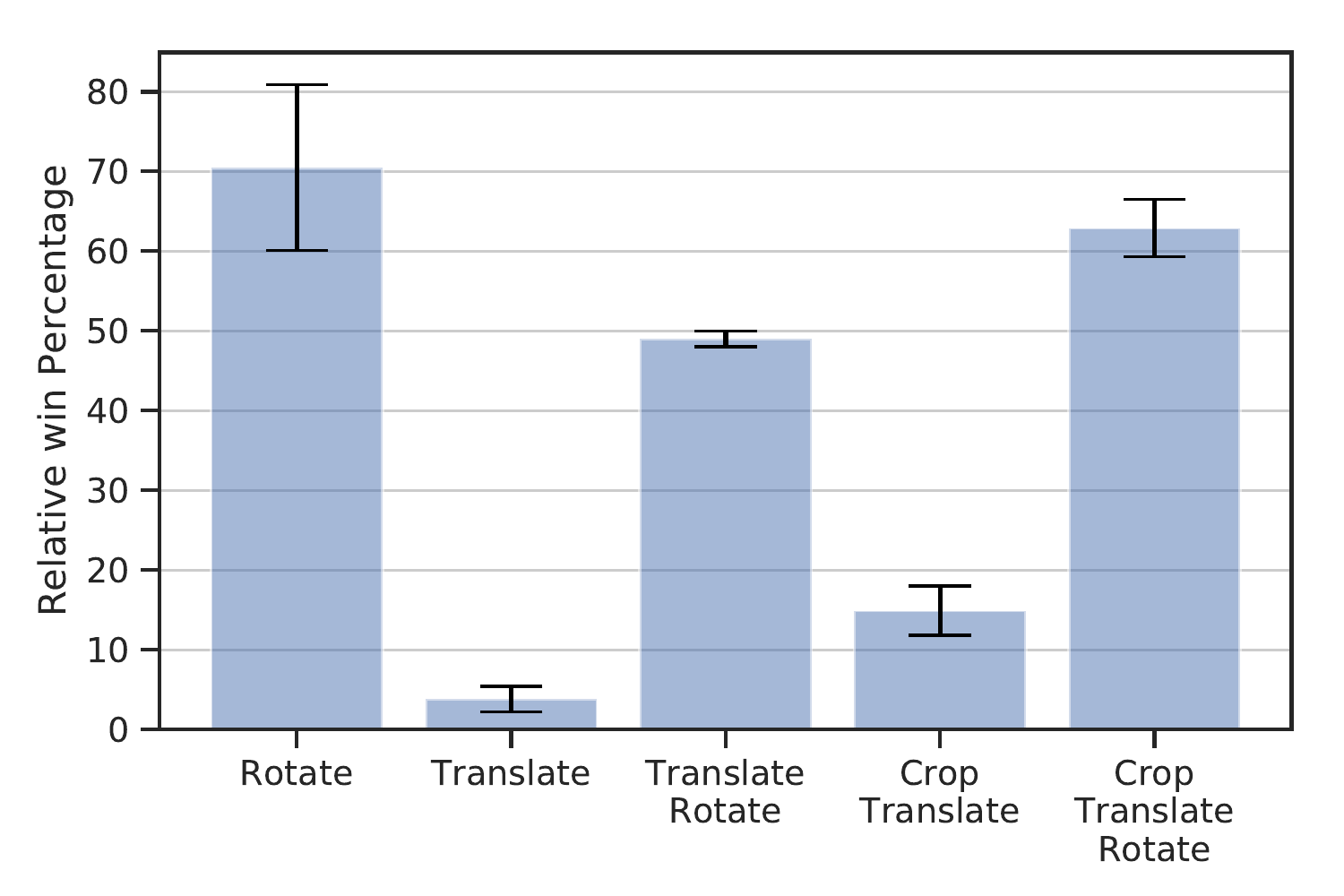}
         \caption{Simple Zelda}
         \label{fig:relative_solarfox}
     \end{subfigure}
     \begin{subfigure}[t]{0.24\linewidth}
         \includegraphics[width=\linewidth]{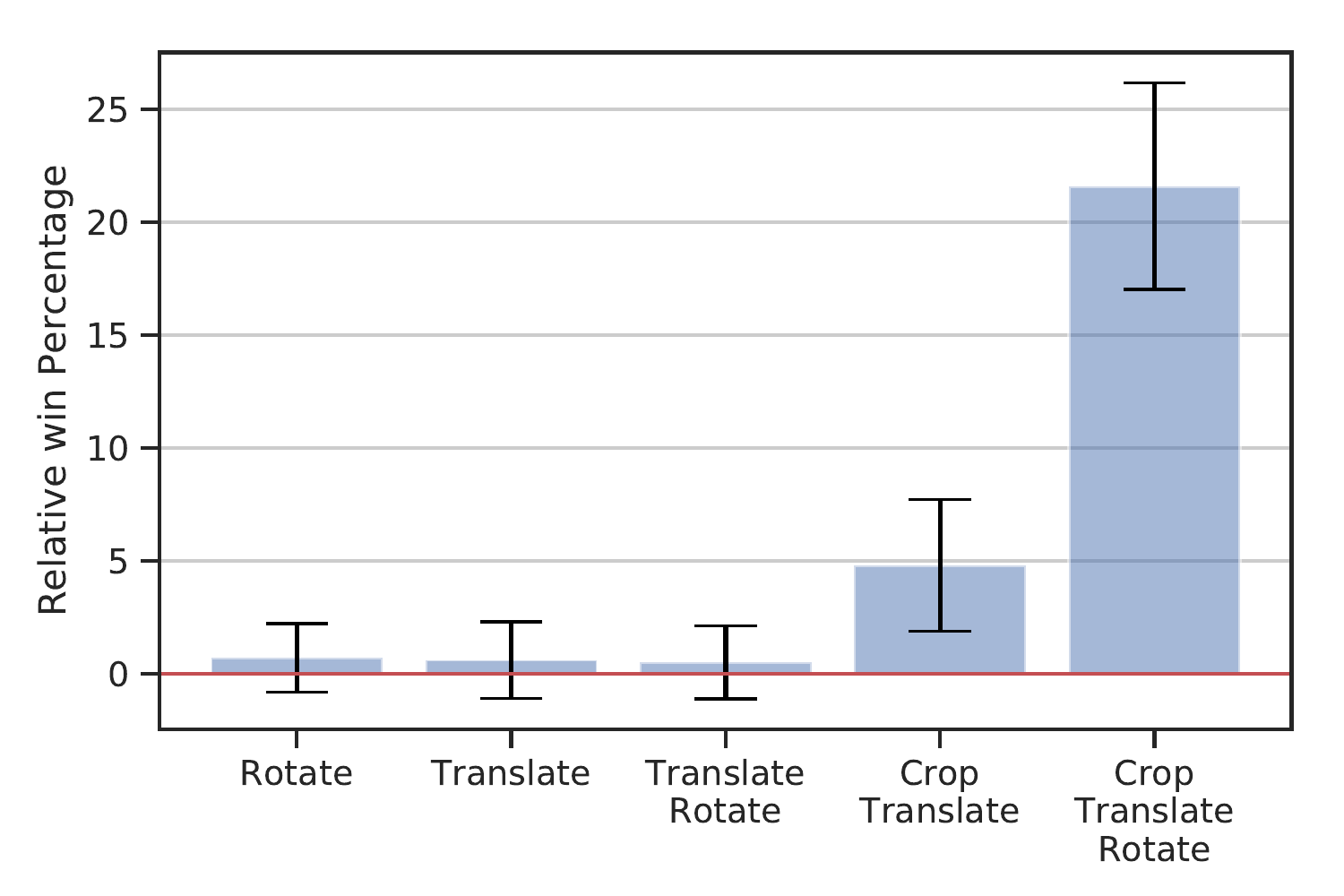}
         \caption{Zelda}
         \label{fig:relative_zelda}
     \end{subfigure}
     \begin{subfigure}[t]{0.24\linewidth}
         \includegraphics[width=\linewidth]{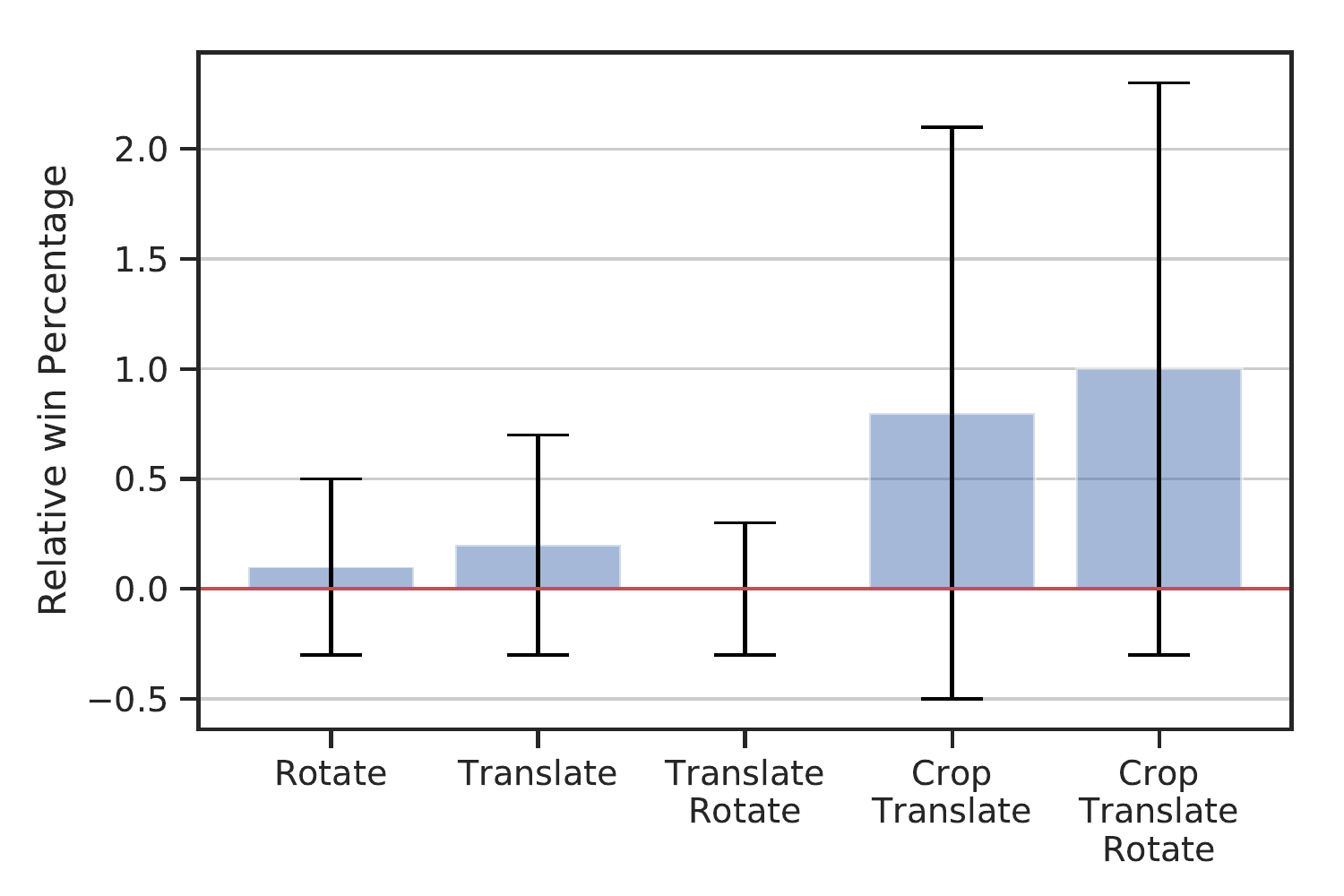}
         \caption{Boulderdash}
         \label{fig:relative_boulderdash}
     \end{subfigure}
     \begin{subfigure}[t]{0.24\linewidth}
         \includegraphics[width=\linewidth]{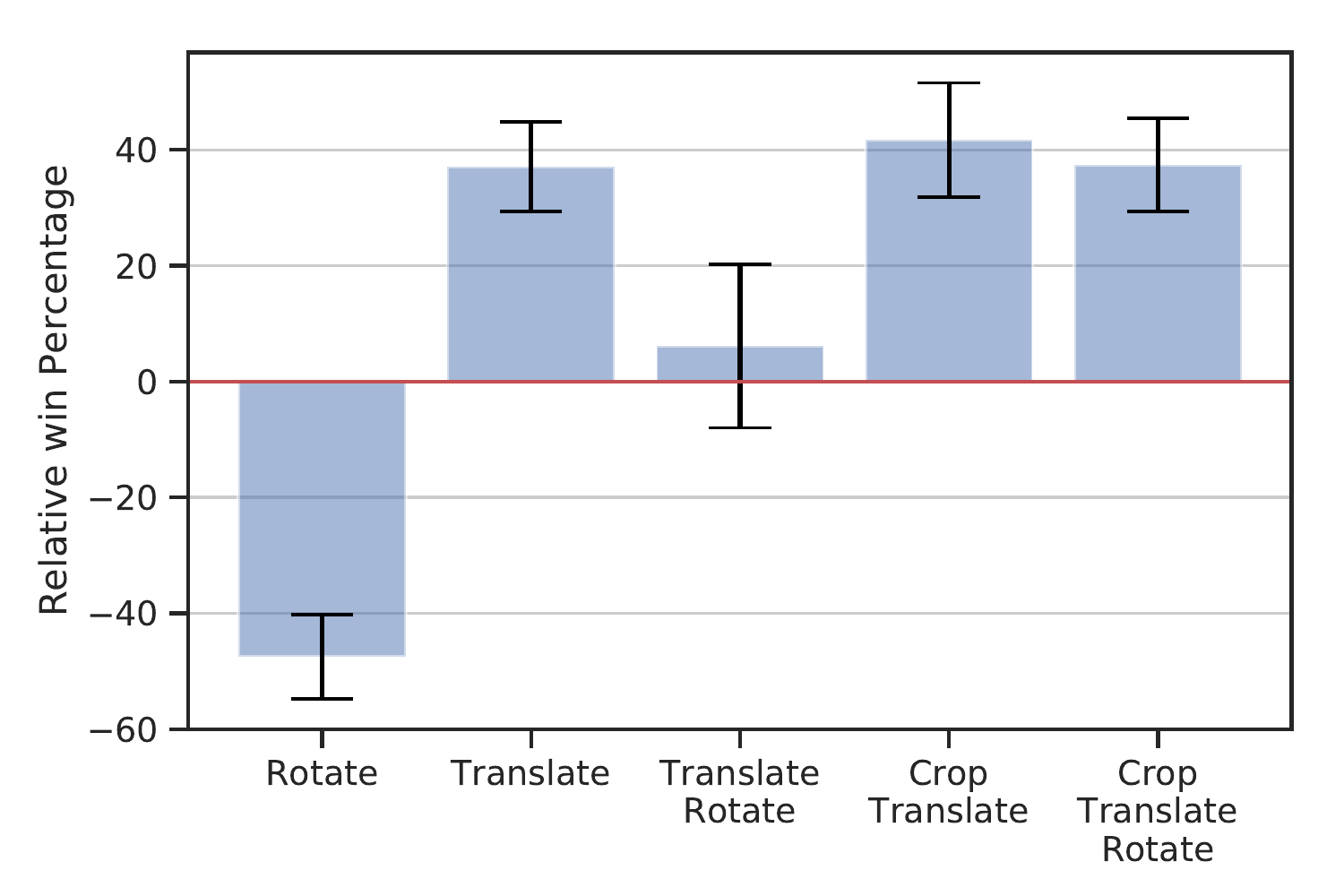}
         \caption{Discrete Solarfox}
         \label{fig:relative_solarfox}
     \end{subfigure}
     \caption{Relative performance of different transformations with respect to the original model performance on the test set}
     \label{fig:relative_performance}
\end{figure*}

For each trained model we test it for 20 times on every level in the training set and test set. Table~\ref{table:results} shows the results from concatenating these data by showing the mean and the standard deviation between the three different models. The low standard deviation shows the model stability during the training process where it achieves almost the same results. Figure~\ref{fig:relative_performance} shows the results of all the algorithms on the test set as a relative performance with respect to the original model performance on the test set. Positive values indicate that this transformation is helpful to the system, while negative values indicate that this transformation is hurting the system, and near-zero values means they are no different.

% We compare all the combinations of the our techniques which end up in being  (remember  approaches, default observation, rotated observation, translated observation, translated and rotated observation, cropped and translated translation and cropped, translated and rotated observation and test each one with or without replacing avatar. We train the agent with each approach 3 times and use the model trained to tested on training level and testing level 1 time. Each level is tested 20 times. We evaluate the performance by calculating the average win rate and its standard deviation. The results are showed in Tabel \ref{table:results}.

% \subsection{Simple Zelda}
In the simple Zelda, all the experiments scored 100\% win rate on the training set which was not surprising as the task is pretty simple and there is no navigation obstacles. Looking on the test set, the cropped, rotated and translated observation achieves the second highest win rate on test levels. Surprisingly, the rotated observation achieves the best win rate on the test set. But the first and second win rate are close due to the high variance in the rotated observation test. Followed by having rotation and translation in the third place. Looking at the rest of the experiments, we find that the agent struggles on the test levels without having the rotation.

% Looking on the test set, we found that having a rotation was enough to perform the best which shows that having rotation can help the agent dramatically in the navigation task. 

%   The agent trained under the dafault observation is struggling on test levels even with avatar replaced. We believe that the rotated observation with avatar replaced outperforms other approaches is mainly because simple zelda does not have random enemies and only have a key and door. There is not many information represented in the observation. And the test levels are identical to training level excepet the location. Therefore, an input contains all the information is reasonable to achieve a high win rate.******. Comparing the approaches under the same cropping and translation setup. The rotation improves the performance on test levels remarkably.

% \subsection{Zelda}
In Zelda, we found that the overall performance on both training set and test set are dropped, especially on the test set. We think that this performance drop is due to the complexity of this game as it has more tasks (navigation through walls and avoiding randomly moving enemies) that it needs to master. Also, the training set size is a lot smaller compared to the simple Zelda experiment. All the trained agents are having similar good performance on the training set with not a huge difference but on the test set the gap is big. The cropped, rotated and translated observation achieves the highest win rate across all approaches. Surprisingly, all the other experiments perform pretty badly especially if it doesn't have cropping. We believe the agent simply overfit to the training levels due to the small training set and couldn't figure a general strategy. On the other hand, when the observation was cropped, the new observation might be more general and more frequently appearing in other levels which helped it to generalize better.

In Boulderdash, the agent struggles to learn to play the game well on the training set which didn't help it to work on the test set with a 0\% win rate. We think this bad performance is because the agent doesn't have any indication about how many diamonds it has collected so far, and it might be impossible to make sure it collects all the diamonds. (On some levels, it is impossible to collect all the diamonds.) This can be noticed from the slightly lower performance of the agent on the training set when it has cropping compared to the rest of the experiments. We think either having a visual indicator or adding memory to the agent might improve its ability to learn and play the game. 

In Discrete Solarfox, the agent learns to play the game pretty good on the training set in most of the cases except with having rotation only. On testing, it is clear that translation is the key component towards generalizing in that game. We think it is due to the nature of that game and the need of relative locations between the avatar and different game objects to perform well in the game. The avatar needs to be far away from the edges of the screen and enemy bullets, while being close to gems to collect them. The case of having only rotation might made it harder for the agent to extract these informations from the running game.

\begin{figure}[ht]
    \centering
    \includegraphics[width=0.6\linewidth]{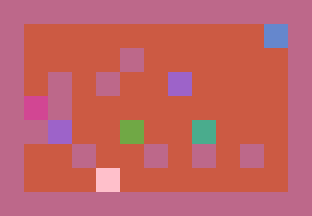}
    \caption{Obvservation with every object replaced, it did not improve generalization.}
    \label{fig:replacing all}
\end{figure}

As mentioned several times, replacing the avatar with a square was shown to always improve testing performance significantly. Extrapolating from this, it seems further removing orientation information by replacing every object in the game with a square would further assist in generalization (see Figure \ref{fig:replacing all}). However, the results are similar to simply replacing the avatar. This is ideal as only replacing the avatar is much easier than replacing every object when only pixel data is available.

\section{Discussion}
Cropping, rotation and translation improve the generalization. However, the win rate is still not promising on more complex problems (Zelda and Boulderdash). We think that could be because of the small training set, the small capacity of our network (3 convolutional layers and 1 fully-connected layer), or the need for memory. From Cobbe's work \cite{cobbe2019leveraging}, larger structures such as IMPALA-CNN \cite{espeholt2018impala} significantly improve generalization comparing to the structure we used in the project which we could adopt for future work. %We might adopt a bigger structure to test out approach in the future work.

The cropping, rotation and translation could be used in many types of games, However, the techniques have some trade-offs and limits. The cropping will throw away global information. This problem is not affecting the performance in Zelda and Solarfox because the actions in these 2 games only affect relative objects with no global effect on the environment. In other games with global effects, cropping might not work. For instance, imagine a scenario where the avatar needs to kill an enemy before it gets stronger, if we use the cropped view, the avatar might not be able to see if there is an enemy to go and kill it before it gets stronger. The rotation technique is based on the assumption that the avatar has a direction it's facing. Therefore, this technique cannot be applied to the games without this property.

The experiments also expose the weakness of the current neural network structure that we are using. The neural network is not always focusing on the area we want it to focus such as the objects in the surrounding. It focuses on tiny object details which was the reason to replacing the avatar in all our experiments. A similar situation also happens in Song et al's work \cite{song2020observational}. The agent focus on the scoreboard instead of the object we want them to focus. By blacking out the scoreboard, the performance on generalization is significantly improved. However, the neural network itself should be able to figure out how to focus on important areas like the selective filtering \cite{broadbent2013perception} in human's visual system and the attention mechanism \cite{vaswani2017attention, wang2018non}. It could be a future research direction on generalization.

As mentioned above, the neural network simply could be viewed as a LSH function. Another interest research direction is combining LSH with feature extraction techniques such as autoencoder to test whether it can achieve similar performance comparing to the regular neural networks.

\section{Conclusion}

This work demonstrates the importance of an agent's perspective when learning. Our three proposed simple changes make a big difference in the policies that the agent learns. This highlights how little is still understood about what causes an agent to learn brittle or robust policies in deep reinforcement learning. This work advances the state-of-the-art for zero-shot generalization as well as formalizes some deep learning tribal knowledge on how to design useful state observations.

The results demonstrate the importance of all three transformations: rotation, translation, and cropping. Giving the agent a narrow, agent-centric view, where it's always facing forward allows it to more accurately learn the effect of each of its actions and the effect of the environment on it. Training on only five levels, it is then able to beat up to $90\%$ of the new levels it had never seen before in a highly stochastic game. That is a huge improvement over what has been possible with so little data.

%We explore the potential of generalization on a small set of training levels. We design three approaches, rotation, translation and cropping, to investigate the affect of input representation on generalization. We test all the combinations of these approaches on two game variants: zelda and simple zelda. Our experiments shed light on why the static third-person views could not generalize and how do we trade off the global information and local information for the agent to generalize. We demonstrate that an ego-centric view with cropping and rotation could improve the generalization on complex tasks such as Zelda, but any other combinations of approaches could not achieve any generalization. While having at least a rotation was sufficient enough to perform well on Simple Zelda. The results indicate that the input representation plays a huge role in generalization which validates our hypothesis that deep reinforcement learning could generalize through a more agent-centric view.

For future work, we would like to continue to test these generalization effects on different games. We would like to continue and improve our understanding of the effects of each of these transformations. It is also important to test these techniques on games where the actions have larger effects on the game state and/or the global game information has more influence on the win rate than local information does. This would give more insight into the efficacy of these techniques in a more diverse set of situations. Finally, since the data-augmentation is a side-effect of our techniques, we would like to apply random data augmentation techniques. Instead of hard coding the augmentation techniques, we could adopt a similar model to Ha and Schmidhuber's World Models
~\cite{ha2018recurrent}. Specifically, we could apply random data augmentations to the input of the vision model so that the model could learn a better representation, similar to the recently published Network Randomization\cite{Lee2020Network}. All of these refinements should help everyone's understanding of some of the real factors that allow for robust policies in some environments and impossible situations in others.

\section*{Acknowledgments}
Ahmed Khalifa acknowledges the financial support from NSF award number 1717324 (``RI: Small: General Intelligence through Algorithm Invention and Selection.''). The authors thank Per Josefsen and Nicola Zaltron, who created for the 45 human-designed Zelda levels.

\bibliographystyle{IEEEtranN}
\bibliography{IEEEabrv,IEEEbiblo}

\end{document}